\documentclass[conference]{IEEEtran}
\IEEEoverridecommandlockouts
\usepackage{caption}
\usepackage{subcaption}
\ifCLASSINFOpdf
\else
   \usepackage[dvips]{graphicx}
\fi
\usepackage{url}
\usepackage{amsmath}
\usepackage{amsthm}
\usepackage{amssymb}
\usepackage[ruled,vlined]{algorithm2e}
\usepackage{breqn}
\newtheorem{theorem}{Theorem}

\def\BibTeX{{\rm B\kern-.05em{\sc i\kern-.025em b}\kern-.08em T\kern-.1667em\lower.7ex\hbox{E}\kern-.125emX}}

\usepackage{graphicx,color}

\DeclareMathOperator*{\argmin}{arg\,min}
\usepackage{textcomp}
\makeatletter
\newcommand{\distas}[1]{\mathbin{\overset{#1}{\kern\z@\sim}}}%
\newsavebox{\mybox}\newsavebox{\mysim}
\newcommand{\distras}[1]{%
  \savebox{\mybox}{\hbox{\kern3pt$\scriptstyle#1$\kern3pt}}%
  \savebox{\mysim}{\hbox{$\sim$}}%
  \mathbin{\overset{#1}{\kern\z@\resizebox{\wd\mybox}{\ht\mysim}{$\sim$}}}%
}

\usepackage{cite}

\title{Federated Smoothing ADMM for Localization}
\author{Reza Mirzaeifard$^{\star}$, Ashkan Moradi$^\S$, Masahiro Yukawa$^\dagger$, Stefan Werner$^{\star}$ \thanks{This work was supported by the Research Council of Norway.}\\

$^{\star}$Dept. of Electronic Systems, Norwegian University of Science and Technology-NTNU, Norway 
\\$^\S$Department of Circulation and Medical Imaging,  Norwegian University of Science and Technology, Norway \\$^\dagger$ Department of Electronics and Electrical Engineering, Keio University, Yokohama, Japan\\
E-mails: \{reza.mirzaeifard,  ashkan.moradi, stefan.werner\}@ntnu.no,
yukawa@elec.keio.ac.jp}

\begin{document}

\maketitle
\maketitle
\maketitle
\begin{abstract}
This paper addresses the challenge of localization in federated settings, which are characterized by distributed data, non-convexity, and non-smoothness. To tackle the scalability and outlier issues inherent in such environments, we propose a robust algorithm that employs an $\ell_1$-norm formulation within a novel federated ADMM framework. This approach addresses the problem by integrating an iterative smooth approximation for the total variation consensus term and employing a Moreau envelope approximation for the convex function that appears in a subtracted form. This transformation ensures that the problem is smooth and weakly convex in each iteration, which results in enhanced computational efficiency and improved estimation accuracy. The proposed algorithm supports asynchronous updates and multiple client updates per iteration, which ensures its adaptability to real-world federated systems. To validate the reliability of the proposed algorithm, we show that the method converges to a stationary point, and numerical simulations highlight its superior performance in convergence speed and outlier resilience compared to existing state-of-the-art localization methods.

\end{abstract}


\section{Introduction}
The rapid expansion of the Internet of Things (IoT) and cyber-physical systems has dramatically increased the amount of data generated by edge devices. Traditional centralized machine learning methods require aggregating this data on a central server, leading to significant privacy concerns and high communication bandwidth costs \cite{yin2021comprehensive}. Federated learning (FL) has emerged as a promising solution to these challenges by enabling decentralized and collaborative model training without requiring clients to share their local data \cite{zhang2024byzantine,zhao2022participant}. 

Despite its advantages, FL faces critical challenges in real-world applications. Outliers in client data can significantly distort the training process, making robust formulations necessary \cite{zhang2024byzantine,zhao2022participant,pillutla2022robust}. Unlike traditional objectives, robust formulations often involve non-convex and non-smooth functions, which may not even be weakly convex \cite{mirzaeifard2023robust}. Many such formulations can be expressed as difference of convex (DC) functions, which are known to be challenging to optimize. Furthermore, FL systems must contend with device heterogeneity, as participating devices vary in computational resources and communication capabilities \cite{pfeiffer2023federated}. This heterogeneity necessitates algorithms that support asynchronous updates and multiple updates per iteration to improve efficiency and reduce communication overhead. However, although recent advances have extended FL beyond smooth and convex objectives, many existing methods still struggle to handle the complexities of robust formulations and device heterogeneity effectively \cite{chen2020asynchronous}. 
This limitation highlights the need for new approaches.

Localization is a fundamental problem in IoT and cyber-physical systems that inherently faces these federated learning challenges, including distributed data, outliers, and device heterogeneity. Localization estimates the position of an object or event in an environment and is widely used in real-world applications\cite{kaiwartya2018geometry}. Common localization techniques include time-of-arrival (ToA) \cite{chan2006time}, received signal strength (RSS) \cite{yin2017received}, time-difference-of-arrival \cite{gustafsson2003positioning}, and angle-of-arrival  \cite{xu2017optimal}. ToA and RSS are distance-based localization methods that leverage existing communication infrastructure to estimate distances between devices, making them particularly suitable for IoT applications \cite{luke2017simple}. While simple and practical, these methods are highly susceptible to errors caused by outliers in the data. Robust localization methods mitigate this issue by minimizing objective functions that penalize large errors, such as the $\ell_1$-norm. These problems are inherently DC optimization tasks, as their objectives can often be expressed as the difference of two convex functions.

Existing methods for addressing localization problems include projection techniques~\cite{jia2010set}, recursive weighted least squares~\cite{wang2014decentralized}, and ADMM-based methods~\cite{zhang2019sensor}. While these approaches are effective in centralized or offline scenarios, they face significant limitations in FL settings. For example, EL-ADMM~\cite{zhang2019sensor} was developed to solve non-convex localization problems without relying on convex relaxation techniques, but it struggles with handling outliers and lacks strong convergence guarantees. Similarly, the distributed subgradient method for robust localization (DSRL) \cite{mirzaeifard2023robust} suffers from slow convergence and inefficiency to support asynchronous and multiple updates. The inherent DC structure of the robust localization problem, coupled with the complexities of asynchronous updates and device heterogeneity in FL systems, makes it a particularly challenging optimization task. Addressing these challenges requires novel FL-based algorithms designed specifically for DC optimization.

To address these challenges, this paper proposes a novel algorithm for robust localization in federated settings, characterized by distributed data, non-convexity, and non-smoothness. The algorithm employs an $\ell_1$-norm formulation within a federated ADMM framework, incorporating an iterative smooth approximation for the total variation consensus term and a Moreau envelope approximation for the convex function that is subtracted in the problem formulation. This transformation ensures that the problem becomes smooth and weakly convex at each iteration, enhancing computational efficiency and estimation accuracy. The algorithm also supports asynchronous updates and multiple client updates per iteration, making it highly adaptable to real-world federated systems. We present a theoretical analysis demonstrating the convergence of the proposed method to a stationary point. Furthermore, numerical simulations reveal that the proposed algorithm surpasses state-of-the-art localization methods by achieving faster convergence and improving resilience to outlier data.

\section{Preliminaries}
We consider the problem of localization in a federated setting with $L$ sensors, where the location of the $l$th sensor is denoted by $\mathbf{a}_l$ for $l = 1, 2, \dots, L$, and the set of all sensor locations is represented as $A = \{\mathbf{a}_1, \mathbf{a}_2, \dots, \mathbf{a}_L\}$. The unknown source is located at $\mathbf{x} \in \mathbb{R}^n$, which must be estimated using range measurements between the source and sensors. 
The observed range $d_l > 0$ between the source and sensor $\mathbf{a}_l$ is modeled as  
\begin{equation}\label{eq1}
    d_l = \|\mathbf{x}-\mathbf{a}_l\|+\epsilon_l, \quad   l = 1,2,\dots,L,
\end{equation}
where $\epsilon_l$ represents measurement noise. In a federated learning setting, each sensor communicates with a central trusted server; however, the exact measurements are restricted from being shared due to privacy and communication constraints.

A common approach for estimating $\mathbf{x}$ is to assume that the measurement noise follows an independent and identically distributed (i.i.d.) Gaussian distribution, leading to the following maximum likelihood estimation (MLE) problem \cite{luke2017simple}:  
\begin{equation}\label{eq2}
\hat{\mathbf{x}}=\argmin_{\mathbf{x}} \frac{1}{L} \sum_{l=1}^{L} \left( \|\mathbf{x}-\mathbf{a}_l\|-d_l\right)^2.
\end{equation}
In real-world scenarios, however, range measurements are often affected by outliers due to sensor failures, multipath effects, or non-line-of-sight conditions. The squared loss function in \eqref{eq2} amplifies the influence of these outliers, making the estimation highly sensitive to outlier measurements.  

To provide a more robust estimation, we can replace the loss function in \eqref{eq2} with an $\ell_1$-norm loss \cite{luke2017simple}:  
\begin{equation}\label{eq3}
\hat{\mathbf{x}}=\argmin_{\mathbf{x}}  \frac{1}{L} \sum_{l=1}^{L} |\|\mathbf{x}-\mathbf{a}_l\|-d_l| = \argmin_{\mathbf{x}}   \sum_{l=1}^{L} f_l(\mathbf{x}),
\end{equation}
where the local objective function at sensor $l$ is  
\begin{equation}
f_l(\mathbf{x})= \frac{1}{L}|\|\mathbf{x}-\mathbf{a}_l\|-d_l|.
\end{equation}
This formulation reduces sensitivity to large errors and makes the solution more robust to outliers. However, it introduces new challenges by transforming the objective function into a non-smooth, non-convex, and even non-weakly convex form, making the optimization problem significantly more difficult. 

ADMM seems like a natural choice for solving \eqref{eq3} in a federated setting due to its advantages in handling non-convex and non-smooth problems in practice \cite{wang2019global,mirzaeifard2023smoothing,mirzaeifard2025federated}. In order to apply conventional ADMM, one might consider rewriting the problem \eqref{eq3} in the following form:
\begin{alignat}{2}\label{eq4}
\nonumber &\min_{\{\mathbf{x}_l\}_{l=1}^{L},\mathbf{w}} &\quad& \sum_{l=1}^{L} f_l(\mathbf{x}_l) \\
&\text{subject to} & &  \mathbf{x}_l = \mathbf{w}, \quad \forall l \in \{1,\dots,L\},
\end{alignat}
where each sensor maintains a local copy $\mathbf{x}_l$ while enforcing consensus through a global variable $\mathbf{w}$. Since $f_l(\mathbf{x})$ is non-convex and non-smooth, ADMM cannot be applied directly, as it requires at least weak convexity \cite{wang2019global, mirzaeifard2023smoothing}. The lack of weak convexity and smoothness also prevents the use of a closed-form proximal update step or a linearized proximal function, which is essential for the iterative structure of ADMM.  

To address these limitations, we first rewrite $f_l(\mathbf{x})$ as a difference of convex (DC) function: 
\begin{equation}
    f_l(\mathbf{x}) = \Gamma_l(\mathbf{x}) - \phi_l(\mathbf{x}),
\end{equation}
which allows partial convexification of the problem with  
$\Gamma_l(\mathbf{x}) = 2\max(0, \|\mathbf{x}-\mathbf{a}_l\|-d_l)$ and $ \phi_l(\mathbf{x}) = \|\mathbf{x}-\mathbf{a}_l\|-d_l$. This decomposition allows us to rewrite the problem as  
\begin{alignat}{2}\label{eq5}
\nonumber &\min_{\{\mathbf{x}_l\}_{l=1}^{L},\mathbf{w}} &\quad& \sum_{l=1}^{L} \Gamma_l(\mathbf{x}_l) - \phi_l(\mathbf{z}_l) \\
&\text{subject to} & & \mathbf{x}_l = \mathbf{z}_l, \quad \mathbf{w} = \mathbf{x}_l\quad \forall l \in \{1,\dots,L\}.
\end{alignat}
This formulation ensures that $\Gamma_l(\mathbf{x})$ has a closed-form proximal function, but $\phi_l(\mathbf{x})$ remains non-smooth and non-convex, preventing direct application of ADMM.  

We now take care of the constraint problem by introducing a relaxation. The strict consensus constraint $\mathbf{x}_l = \mathbf{w}$ necessitates the need for a smooth function when we update each $\mathbf{x}_l$. To mitigate this issue, we replace the hard constraint with a total variation norm, which allows a soft consensus between the local variables while still preserving consistency. The modified problem is reformulated as  
\begin{alignat}{2}\label{eq6}
\nonumber &\min_{\{\mathbf{x}_l,\mathbf{z}_l,\mathbf{q}_l,\mathbf{g}_l\}_{l=1}^{L},\mathbf{w}} &\quad& \sum_{l=1}^{L} \Gamma_l(\mathbf{x}_l)-\phi_l(\mathbf{z}_l)+\omega\|\mathbf{q}_l-\mathbf{g}_l\|_1  \\
&\text{subject to} & & \mathbf{x}_l = \mathbf{z}_l, \quad \forall l \in \{1,\dots,L\}, \\
& & & \mathbf{g}_l = \mathbf{x}_l, \quad \mathbf{q}_l = \mathbf{w}, \forall l \in \{1,\dots,L\}.
\end{alignat}
Since the objective function has bounded gradients, the total variation norm with a suitable $\omega$ does not interfere with the consensus requirement while still allowing flexibility in optimization \cite{mirzaeifard2024decentralized}.  

   After this relaxation, we still need to handle the non-smoothness of the total variation norm regularization term and the remaining nonconvex, non-smooth component $\phi_l(\mathbf{x})$ as they get updated in the second block of ADMM. 
   The next sections discuss the details of the smoothing approach and the convergence guarantees of the proposed method.
\section{Distributed Robust Localization}
To address the challenges of optimizing non-smooth functions using ADMM, smoothing techniques are employed to approximate a function \( g \) with a family of smooth functions \( \tilde{g} \), which preserve essential properties such as continuity and differentiability. These approximations ensure that \( \tilde{g}(\mathbf{x}, \mu) \) converges to \( g(\mathbf{x}) \) as \( \mu \to 0^+\) \cite{chen2012smoothing}, allowing for effective numerical optimization in federated settings.

In our approach, we approximate the $\ell_1$-norm term $\|\mathbf{z}\|_1$ using a smooth function representation for each component $|z_i|$, as introduced in \cite{chen2012smoothing}. Specifically, the smoothing function is defined as:
\begin{equation}\label{eq12}
 f\left(z_i,\mu\right)=
\begin{cases}
|z_i|, &  \mu \leq |z_i| \\
\frac{z_i^2}{2\mu}+\frac{\mu}{2}, &  |z_i| < \mu.
\end{cases}
\end{equation}
This formulation ensures that the smoothed function $h(\mathbf{z},\mu) = \sum_{i=1}^{n} f(z_i,\mu)$ provides an upper bound on $\|\mathbf{z}\|_1$, which converges to the exact $\ell_1$ norm as $\mu$ decreases. To further refine the approximation, we apply a Moreau envelope to each function $\phi_l(\cdot)$, obtaining a smooth lower bound:
\begin{equation}\label{eq_moreau}
e_{\phi_l}(\mathbf{x},\mu) = 
\begin{cases}
\displaystyle \frac{1}{2\mu} \|\mathbf{x} - \mathbf{a}_l\|^2 - d_l & \text{if } \|\mathbf{x} - \mathbf{a}_l\| \leq \mu, \\
\displaystyle \|\mathbf{x} - \mathbf{a}_l\| - \frac{\mu}{2} - d_l & \text{if } \|\mathbf{x} - \mathbf{a}_l\| > \mu.
\end{cases}  
\end{equation} 
Since $-e_{\phi_l}(\mathbf{x},\mu)$ is an upper bound for $-\phi_l(\mathbf{x})$, minimizing it ensures the original function is also minimized. Additionally, $-e_{\phi_l}(\mathbf{x},\mu)$ exhibits $\frac{1}{\mu}$-smoothness with a bounded gradient, ensuring weak convexity (i.e., $-e_{\phi_l}(\cdot,\mu) + \frac{1}{2\mu} \|\cdot\|^2$ is convex).

Leveraging these approximations, we derive the following augmented Lagrangian:
\begin{multline}\label{eq13}
{\bar{\mathcal{L}}}_{\sigma_{\Psi},\sigma_{\xi},\mu_{h},\mu_{\phi}}\left(\mathbf{X},\mathbf{w},\mathbf{Z},\mathbf{Q},\boldsymbol{\Psi},\boldsymbol{\xi},\boldsymbol{\zeta}\right)=\\   \sum_{l=1}^{L} \Bigg( \Gamma_l(\mathbf{x}_l)-e_{\Phi_l}(\mathbf{z}_l,\mu_{\phi}) +\boldsymbol{\Psi}_l^{\top}(\mathbf{x}_l-\mathbf{z}_l) \\ + \frac{\sigma_{\Psi}}{2}\|\mathbf{x}_l-\mathbf{z}_l\|^2_2+
      \omega \hspace{0.5mm} h\left(\mathbf{g}_{l}-\mathbf{q}_{l},\mu_{h}\right)+\frac{\sigma_{\xi}}{2} \|\mathbf{x}_{l}-\mathbf{g}_{l}\|^2_2\\+\frac{\sigma_{\xi}}{2}\|\mathbf{w}-\mathbf{q}_{l}\|^2_2  
+\boldsymbol{\xi}_{l}^{\text{T}} (\mathbf{x}_{l}-\mathbf{g}_{l}) + \boldsymbol{\zeta}_{l}^{\text{T}} (\mathbf{w}-\mathbf{q}_{l})\Bigg)
\end{multline}
where $\mathbf{X}=[\mathbf{x}_1,\cdots,\mathbf{x}_L]$,  $\boldsymbol{\xi}= \{\boldsymbol{\xi}_{l}\}_{l=1}^{L}$, $\boldsymbol{\zeta}= \{\boldsymbol{\zeta}_{l}\}_{l=1}^{L}$, and $\boldsymbol{\Psi}= \{\boldsymbol{\Psi}_{l}\}_{l=1}^{L}$ are dual variables, $\mu_{l,\Psi}$ and $\mu_{l,\xi}$ are smoothing parameters, and $\sigma_{l,\Psi}$ and $\sigma_{l,\xi}$ are penalty parameters.

Each client \( l \) operates independently, maintaining its own update index \( k_l \), which tracks the number of local iterations performed. Asynchronous updates are permitted, allowing clients to perform multiple updates per global iteration \( k_w \). This design accounts for device heterogeneity, making it suitable for federated learning.

To progressively refine the approximations, the parameters \( \mu_g \), \( \mu_h \), \( \sigma_{\psi} \), and \( \sigma_{\xi} \) are updated at each local iteration \( k_l \) following the rules:
\begin{align}\label{eq:up:sm}
\sigma_{\psi}^{(k_l)} = c \sqrt{k_l}, 
\mu_{\phi}^{(k_l)} = \frac{\alpha}{\sqrt{k_l}}, 
\sigma_{\xi}^{(k_l)} = d \sqrt{k_l}, 
\mu_{h}^{(k_l)} = \frac{\beta}{\sqrt{k_l}}
\end{align}
Since smoothing gradually diminishes over iterations, the optimization process progressively recovers the original problem structure, enabling ADMM to operate effectively even in non-smooth and non-convex settings.

To update \( \mathbf{x}_l \), we solve the following optimization problem:
\begin{multline}\label{eq15}
\mathbf{x}^{(k_l)}_l = \arg\min_{\mathbf{x}_l} \;  \Gamma_l(\mathbf{x}_l) + (\boldsymbol{\Psi}_l^{(k_l-1)})^\top (\mathbf{x}_l - \mathbf{z}_l^{(k_l-1)}) \\  
+ \sigma_{\Psi}^{(k_l)} \| \mathbf{x}_l - \mathbf{z}_l^{(k_l-1)} \|^2 
+ \frac{\sigma_{\xi}^{(k_l-1)}}{2} \| \mathbf{x}_l - \mathbf{g}_{l}^{(k_l-1)} \|^2 \\  
+ (\boldsymbol{\xi}_{l}^{(k_l-1)})^\top (\mathbf{x}_l - \mathbf{g}^{(k_l-1)}_{l})
\end{multline}

By grouping the quadratic and linear terms in \eqref{eq15}, we rewrite the optimization problem as:
\begin{equation}\label{eq16}
\mathbf{x}_l = \arg\min_{\mathbf{x}_l} \; \Gamma_l(\mathbf{x}_l) + \frac{\upsilon}{2} \left\| \mathbf{x}_l - \mathbf{p} \right\|^2 = \text{Prox}_{\Gamma_l}(\mathbf{p};\frac{1}{\upsilon}),
\end{equation}
where 
\begin{equation}\label{eq17}
    \upsilon =  \sigma_{\Psi}^{(k_l)} + \sigma_{\xi}^{(k_l)},
\end{equation}
\begin{equation}\label{eq18}
  \mathbf{p} = -\frac{1}{\upsilon} \bigg(\boldsymbol{\Psi}_l^{(k_l-1)} + \boldsymbol{\xi}_{l}^{(k_l-1)} -  \sigma_{\Psi}^{(k_l)} \mathbf{z}_l^{(k_l-1)} - \sigma_{\xi}^{(k_l)}  \mathbf{g}_{l}^{(k_l-1)} \bigg)
\end{equation}
The proximal operator of \( \Gamma_l \) is given by:
\begin{multline}\label{eq19}
\text{Prox}_{\Gamma_l}(\mathbf{p};\lambda) = \\  
\begin{cases}
\mathbf{p}, & \text{if } \|\mathbf{p} - \mathbf{a}_l\| \leq d_l, \\
\mathbf{a}_l + d_l \dfrac{ \mathbf{p} - \mathbf{a}_l }{ \|\mathbf{p} - \mathbf{a}_l\| }, & \text{if } d_l < \|\mathbf{p} - \mathbf{a}_l\| \leq d_l + 2\lambda, \\
\mathbf{p} - 2\lambda \dfrac{ \mathbf{p} - \mathbf{a}_l }{ \|\mathbf{p} - \mathbf{a}_l\| }, & \text{if } \|\mathbf{p} - \mathbf{a}_l\| > d_l + 2\lambda.
\end{cases}
\end{multline}

To update \( \mathbf{z}_l \), we solve the following optimization problem:
\begin{multline}\label{eq20}
\mathbf{z}_l = \arg\min_{\mathbf{z}_l} \; -e_{\phi}(\mathbf{z}_l,\mu_\phi^{(k_l)}) + (\boldsymbol{\Psi}_l^{(k_l-1)})^\top (\mathbf{x}_l^{(k_l)} - \mathbf{z}_l) \\ 
+ \frac{\sigma_{\Psi}^{(k_l)}}{2} \| \mathbf{x}_l^{(k_l)} - \mathbf{z}_l \|^2
\end{multline}
By rearranging and completing the square, we rewrite \eqref{eq20} as:
\begin{equation}\label{eq21}
\mathbf{z}_l =\arg\min_{\mathbf{z}_l} \; -e_{\phi}(\mathbf{z}_l,\mu_\phi^{(k_l)}) +   
\frac{\sigma_{\Psi}^{(k_l)}}{2} 
\left\| \mathbf{z}_l -  \mathbf{x}_l^{(k_l)} + \frac{\boldsymbol{\Psi}_l^{(k_l-1)}}{ \sigma_{\Psi}^{(k_l)}}  \right\|^2
\end{equation}
Thus, the update reduces to computing the proximal operator:
\begin{equation}\label{eq22}
\mathbf{z}_l = \text{Prox}_{-e_{\phi}(\cdot,\mu_\phi^{(k_l)})}(\mathbf{c};\alpha),
\end{equation}
where \( \alpha = \dfrac{1}{ \sigma_{\Psi}^{(k_l)}} \) and
\(
\mathbf{c} = \mathbf{x}_l^{(k_l)} - \frac{\boldsymbol{\Psi}_l^{(k_l-1)}}{\sigma_{\Psi}^{(k_l)}}
\). The $\text{Prox}_{-e_{\phi}(\cdot,\mu)}(\cdot;\frac{1}{\sigma})$ is a single-valued operator for every $\sigma > \frac{1}{\mu}$, because the function 
\(
-\frac{e_\phi(\cdot,\mu)}{\sigma}
\)
is weakly convex with modulus $\frac{1}{\mu\sigma}$ \cite{yukawa2024monotone}.
 The proximal operator is given by:
\begin{multline}\label{eq23}
\text{Prox}_{-e_{\phi_l}(\cdot,\mu_\phi)}(\mathbf{c};\alpha)=\mathbf{a}_l +\\ 
\begin{cases}
 \frac{\mu_\phi}{\mu_\phi-\alpha} \,(\mathbf{c}-\mathbf{a}_l), & \text{if } \| \mathbf{c} - \mathbf{a}_l \| \leq \mu_\phi - \alpha  \\
\frac{\|\mathbf{c}-\mathbf{a}_l\|+\alpha}{\|\mathbf{c}-\mathbf{a}_l\|} \,(\mathbf{c}-\mathbf{a}_l), & \text{if } \| \mathbf{c} - \mathbf{a}_l \| > \mu_\phi - \alpha
\end{cases}
\end{multline}
For the updates of both \( \mathbf{g}_{l} \) and \( \mathbf{q}_{l} \) in parallel, we have:
\begin{multline}\label{eq24}
  \begin{bmatrix}
 \mathbf{g}_{l}^{(k_l)}\\
    \mathbf{q}_{l}^{(k_l)}
\end{bmatrix} =\argmin_{\mathbf{g}_{l},\mathbf{q}_{l}}  \omega \sum_{p=1}^{n} f({g}_{l,p}-{q}_{l,p},\mu^{\left(k_l\right)})-\frac{\sigma_{\xi}^{(k_l)}}{2}\times \\ 
\phantom{=}\Bigg(   \left\|\mathbf{g}_{l}-\mathbf{x}_{l}^{(k_l)}+\frac{\boldsymbol{\xi}_{l}^{(k_l-1)}}{\sigma_{\xi}^{(k_l)}}\right\|^2_2+ \left\|\mathbf{q}_{l}-\mathbf{w}^{(k_w)}-\frac{\boldsymbol{\zeta}_{l}^{(k_l-1)}}{\sigma_{\xi}^{(k_l)}}\right\|^2_2\Bigg)  
\end{multline}
Following the simplification provided in \cite{hallac2017network}, we get:
\begin{dmath}\label{up:g}
     \begin{bmatrix}
    \mathbf{g}_{l}^{(k_l)}\\
    \mathbf{q}_{l}^{(k_l)}
\end{bmatrix}= 
\frac{1}{2}\begin{bmatrix}\mathbf{x}_{l}^{(k_l)}+\frac{\boldsymbol{\xi}_{l}^{(k_l-1)}}{\sigma_{\xi}^{(k_l)}}+\mathbf{w}^{(k_w)}+\frac{\boldsymbol{\zeta}_{l}^{(k_l-1)}}{\sigma_{\xi}^{(k_l)}}\\
\mathbf{x}_{l}^{(k_l)}+\frac{\boldsymbol{\xi}_{l}^{(k_l-1)}}{\sigma_{\xi}^{(k_l)}}+\mathbf{w}^{(k_w)}+\frac{\boldsymbol{\zeta}_{l}^{(k_l-1)}}{\sigma_{\xi}^{(k_l)}}
\end{bmatrix}
+\frac{1}{2}\begin{bmatrix}
 -\mathbf{e}\\
\mathbf{e}
\end{bmatrix}
\end{dmath}
where for each component \( p \in  \{1, \ldots, P+1\} \),
\({e}_p= \text{Prox}_{f(\cdot, \mu_h^{(k_l)})}\left(
 -{x}_{l,p}^{(k_l)} - \frac{{\xi}_{l,p}^{(k_l-1)}}{\sigma_{\xi}^{(k_l)}}
+ {w}_{p}^{(k_w)} + \frac{{\zeta}_{l,p}^{(k_l-1)}}{\sigma_{\xi}^{(k_l)}};
\frac{2\omega}{\sigma_{\xi}^{(k_l)}}
\right)\) as defined in \cite{mirzaeifard2023smoothing}.
Furthermore, the update for each dual variable \( \boldsymbol{\Psi}_l \), \( \boldsymbol{\xi}_l \), and \( \boldsymbol{\zeta}_{l} \) is given by:
\begin{equation}\label{eq28}
\boldsymbol{\Psi}_{l}^{(k_l)}=\boldsymbol{\Psi}^{(k_l-1)}_{l}+\sigma_{\Psi}^{(k_l)}\left(\mathbf{x}^{(k_l)}_l-\mathbf{z}^{(k_l)}_{l}\right)
\end{equation} 
\begin{equation} \label{eq29}
\boldsymbol{\xi}_{l}^{(k_l)}=\boldsymbol{\xi}^{(k_l-1)}_{l}+\sigma_{\xi}^{(k_l)}\left(\mathbf{x}^{(k_l)}-\mathbf{g}_l^{(k_l)}\right)
 \end{equation}
 \begin{equation} \label{eq30}
\boldsymbol{\zeta}_{l}^{(k_l)}=\boldsymbol{\zeta}^{(k_l-1)}_{l}+\sigma_{\xi}^{(k_l)}\left(\mathbf{w}^{(k_w)}-\mathbf{q}_l^{(k_l)}\right)
 \end{equation}
 Accordingly, we set \( k_l^w \) to \( k_l \) to provide updated values for the central client.

After all clients update their local variables, the central server performs the following update step for the global variable \( \mathbf{w} \), aggregating information from all clients:
\begin{equation}\label{Eq.8}
   \mathbf{w}^{(k_w)} =  \frac{\sum_{l=1}^L\sigma_{\xi}^{(k_l^w+1)}\mathbf{q}^{(k_l^w)}_l-\boldsymbol{\zeta}^{(k_l^w)}_l}{\sum_{l=1}^L\sigma_{\xi}^{(k_l^w+1)}},
\end{equation}
where \( k_w \) denotes the iteration of the global update. This update ensures consistency across clients by incorporating latest updates, aligning local and global optimization objectives.

\begin{algorithm}[t]
 \caption{Federated Smoothing with Moreau envelope ADMM (FSMDM)}
 \label{alg:1}
\SetAlgoLined
Initialize $c$, $d$, $\beta$, $K_l$ for each node $l$, $K_z$, $K_w$  and the parameters $\tau$, $\gamma$, and $\lambda$\;
 
  \For{each node $l = 1, \cdots, L$ in parallel}{
   Update $\mu_\phi^{(k_l)}$, $\sigma_\xi^{(k_l)}$, $\mu_h^{(k_l)}$ and $\sigma_\psi^{(k_l)}$ by \eqref{eq:up:sm}\;
   \For{each local iteration $k_l = 1, \cdots, K_l$ (asynchronously)}{
    
    Update $\mathbf{x}_l^{(k_l)}$ by \eqref{eq16}\;

    Update $\mathbf{z}_l^{(k_l)}$ by \eqref{eq22}\;

     Update $\boldsymbol{\psi}_l^{(k_l)}$ by \eqref{eq28}\;

    Receive updated variable from the central node\;
    Update $\mathbf{g}_l^{(k_l)}$ and $\mathbf{q}_l^{(k_l)}$ by \eqref{up:g}\;
    Update  $\boldsymbol{\xi}_l^{(k_l)}$ and $\boldsymbol{\zeta}_l^{(k_l)}$ by \eqref{eq29} and \eqref{eq30} and send  updated variables to  server\;
   }
   
  }
  \For{each global iteration $k_w = 1, \cdots, K_w$}{
  Receive updated variables from all nodes\;
  Update $\mathbf{w}^{(k_w)}$ by \eqref{Eq.8} and 
  send it to all nodes\;}
 
\end{algorithm}

We note that the asynchronicity of local updates does not compromise the validity of the convergence proof. The updates to the dual variables, following an ascent step, occur after the updates of \( \mathbf{g}_l \) and \( \mathbf{q}_l \). This order of updates mitigates the impact of dual variable updates and ensures the convergence of the algorithm while maintaining robustness to noise introduced by smoothing, as the noise impact is bounded. To maintain synchronization, we assume that while the global update for \( \mathbf{w} \) occurs, local updates for \( \mathbf{g}_l \), \( \mathbf{q}_l \), \( \boldsymbol{\xi}_l \), and \( \boldsymbol{\zeta}_l \) are paused. Similarly, during local client updates, the global variable update is temporarily suspended. Furthermore, there exists a constant \( K_a \) such that within every \( K_a \) 
global steps, each client updates its local variables at least once.

\begin{figure*}[t!]
  \centering
  \begin{minipage}{0.27\textwidth}
    \centering
    \includegraphics[width=\linewidth]{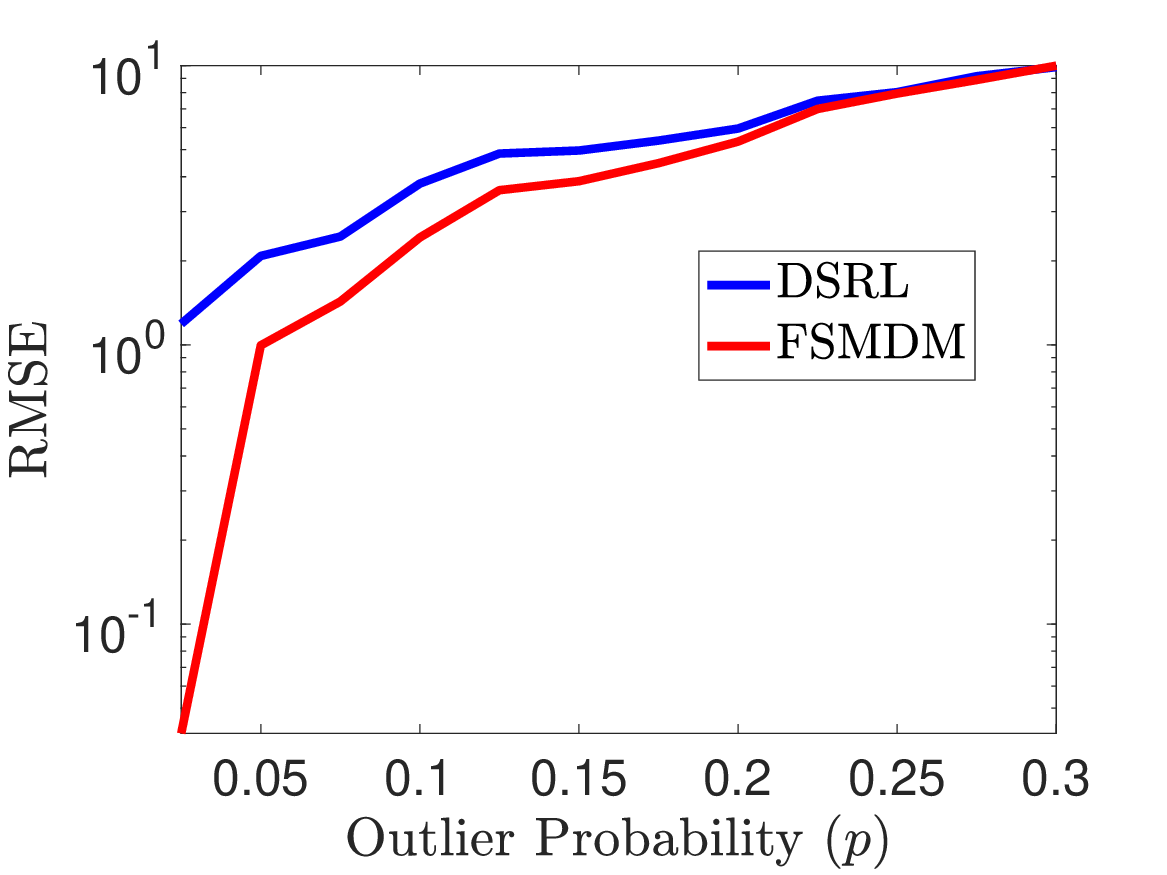}
    \captionof{figure}{RMSE versus outlier probability.}
    \label{fig1}
  \end{minipage}\hfill
  \begin{minipage}{0.27\textwidth}
    \centering
    \includegraphics[width=\linewidth]{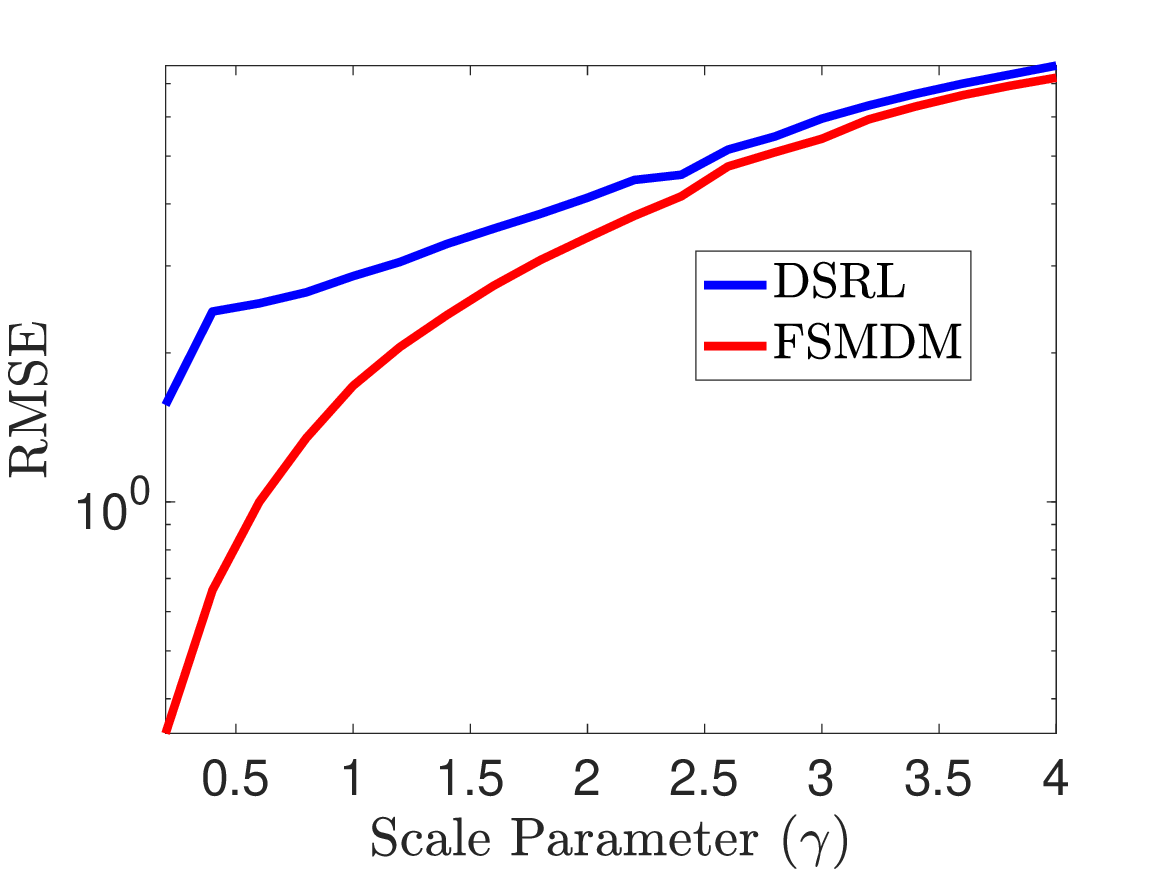}
    \captionof{figure}{RMSE versus scale parameter of Cauchy noise.}
    \label{fig3}
  \end{minipage}\hfill
  \begin{minipage}{0.27\textwidth}
    \centering
    \includegraphics[width=\linewidth]{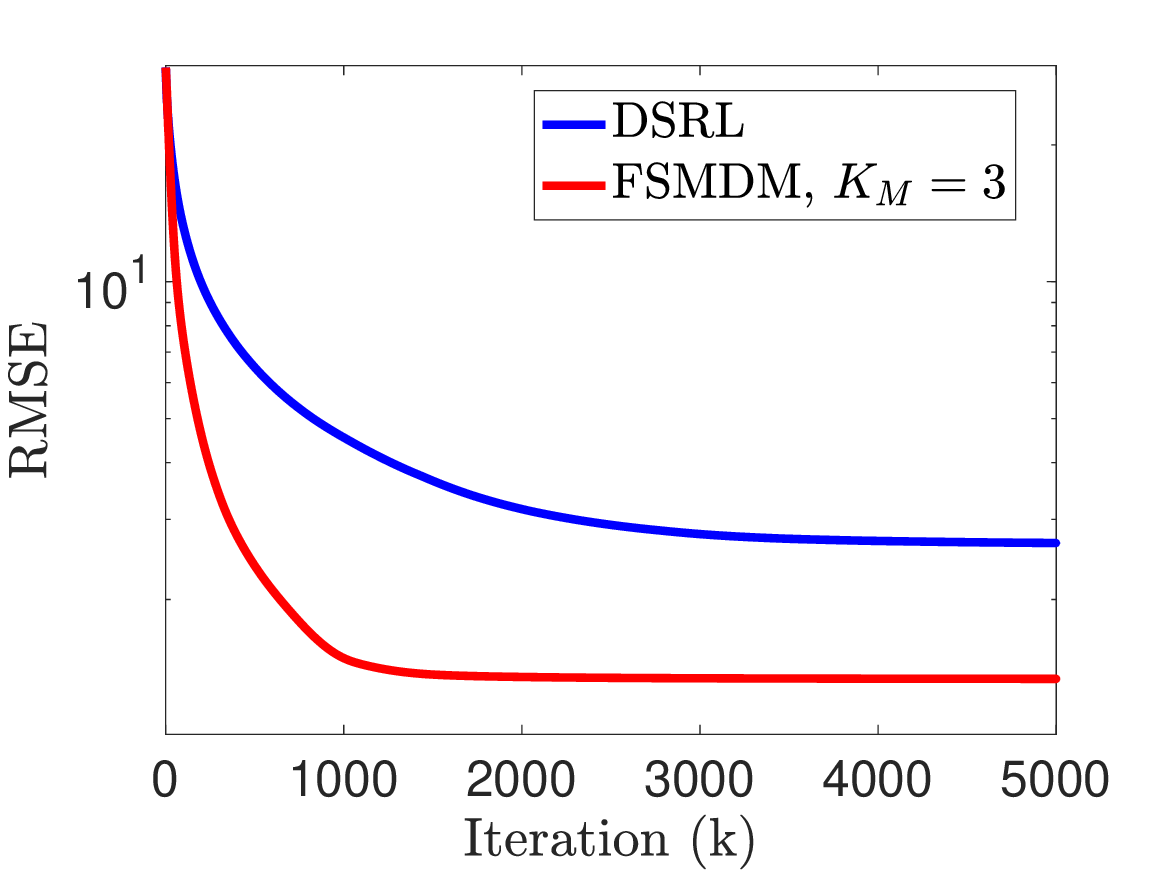}
    \captionof{figure}{RMSE versus iterations.}
    \label{fig4}
  \end{minipage}
\end{figure*}

\begin{theorem}[Global Convergence]\label{theorem1}
Assume that the  parameters $\mu_{\xi}^{(k_l)}$, $\mu_{h}^{(k_l)}$, $\sigma_{\phi}^{(k_l)}$ and $\sigma_{\psi}^{(k_l)}$ are updated according to \eqref{eq:up:sm},  $\beta d \geq \sqrt{80}$ and $\alpha c \geq \sqrt{\frac{3}{2}}$. 
Then, Algorithm \ref{alg:1} converges to a stationary point $\left(\{\mathbf{x}_l^{*}, \boldsymbol{\Psi}_l^{*}, \boldsymbol{\xi}_l^{*}, \boldsymbol{\zeta}_l^{*}\}_{l=1}^{L}, \mathbf{w}^{*}\right)$ satisfying the following KKT conditions:
\begin{subequations}
\begin{align}
& \boldsymbol{\xi}_l^{*} \in \Gamma_l(\mathbf{x}_l^{*})-\phi_l(\mathbf{x}_l^{*})\\
&\sum_{l=1}^L \boldsymbol{\zeta}_l^{*} = \mathbf{0} \\
&\mathbf{w}^{*} = \mathbf{x}_l^{*}, \quad \forall l \in \{1, \ldots, L\},
\end{align}
\end{subequations}
where \(\boldsymbol{\zeta}_l^{*} = -\boldsymbol{\xi}_l^{*}\) for each \( l \in \{1,\cdots,L\} \).
\end{theorem}

\section{Simulation Results}
In this section, we evaluate the performance of the proposed federated smoothing ADMM with Moreau envelope (FSMDM) through simulations and compare it with the federated version of DSLR  \cite{mirzaeifard2023robust}. The sensor network $\mathcal{G}$ consists of $L=21$ clients, which are uniformly and randomly positioned within a cubic region of $[-30,30]^{3}$. The target location is also uniformly and randomly distributed within the same region. To assess accuracy, we use the root-mean-square error (RMSE) as the performance metric, defined as \(\text{RMSE} = \sqrt{\frac{\sum_{i=1}^L||\mathbf{\hat{w}}_i-\mathbf{x}||_2^2}{L}}.\)
The results are obtained by averaging over 1000 independent trials. The following parameters are used in the simulations: $c=\frac{1}{100\sqrt{2}}$, $d=\frac{\sqrt{3}}{\sqrt{500}}$, $\alpha=100\sqrt{3}$, and $\beta=100\sqrt{3}$.

In the first scenario, we analyze the robustness of the algorithms in the presence of outliers. We introduce outliers with varying probability, denoted as $p$, ranging from $0.025$ to $0.3$ in increments of $0.025$. The outliers, represented by $\epsilon_i$, follow a uniform distribution over the interval $[0, 60\sqrt{3}]$, as modeled by:
$
y_i=
\begin{cases}
\epsilon_i, & \text{if } u_i <p, \\
\|\mathbf{x}-\mathbf{a}_i\|, & \text{otherwise},
\end{cases}
$
where $u_i$ is sampled from a uniform distribution on the interval $[0,1]$. As shown in Figure \ref{fig1}, FSMDM consistently outperforms DSLR across all outlier probabilities, demonstrating its superior robustness.

In the second scenario, we examine the accuracy of the algorithms when the measurement noise follows a Cauchy distribution. The measurements are modeled as in \eqref{eq1}, with $\epsilon_i$ independently drawn from a Cauchy distribution with scale parameter $\gamma$, i.e., $\epsilon_i \sim \text{Cauchy}(0, \gamma)$. Simulations are conducted for $\gamma$ values ranging from $0.2$ to $4$ in increments of $0.2$. Figure \ref{fig3} demonstrates that FSMDM achieves lower RMSE compared to DSLR across all tested scale parameters, highlighting its robustness to heavy-tailed noise. Moreover, when $\gamma=1$, Figure \ref{fig4} shows that FSMDM exhibits a faster convergence rate than DSLR, particularly when three local updates are performed per global iteration in FSMDM.
\section{Conclusion}
This paper introduces a federated ADMM-based algorithm for robust localization, integrating smoothing techniques to tackle non-convexity and non-smoothness. By leveraging the Moreau envelope for convex functions that are subtracted in the problem formulation, and a smoothed total variation norm, the method ensures computational efficiency and improved convergence. The algorithm is designed to support asynchronous updates and multiple local updates per global iteration, making it well-suited for real-world federated learning environments. Simulations show FSMDM outperforms state-of-the-art methods like DSLR, achieving lower RMSE, faster convergence, and robustness to outliers and heavy-tailed noise, making it well-suited for distributed localization tasks.
\bibliographystyle{IEEEtran}
\bibliography{strings,refs}

\end{document}